\documentclass[runningheads]{llncs}
\usepackage{graphicx}
\usepackage{threeparttable}
\begin{document}
\title{Semantic similarity estimation for domain specific data using BERT and other techniques
\thanks{This is a preprint version of an article accepted for publication in the proceedings of Machine Learning and Data Mining 2019.}}

%
\titlerunning{Semantic similarity estimation for domain specific data }
%
\author{R. Prashanth}
\authorrunning{R. Prashanth}
%
\institute{Fidelity Business Services India Pvt. Ltd., \\ Embassy Golf Links Business park, Bangalore 560 071, India\\
\email{prashanth.r@fmr.com}}

\maketitle              
\begin{abstract}
Estimation of semantic similarity is an important research problem both in natural language processing and the natural language understanding, and that has tremendous application on various downstream tasks such as question answering, semantic search, information retrieval, document clustering, word-sense disambiguation and machine translation. In this work, we carry out the estimation of semantic similarity using different state-of-the-art techniques including the USE (Universal Sentence Encoder), InferSent and the most recent BERT, or Bidirectional Encoder Representations from Transformers, models. We use two question pairs datasets for the analysis, one is a domain specific in-house dataset and the other is a public dataset which is the Quora's question pairs dataset. We observe that the BERT model gave much superior performance as compared to the other methods. This should be because of the fine-tuning procedure that is involved in its training process, allowing it to learn patterns based on the training data that is used. This works demonstrates the applicability of BERT on domain specific datasets. We infer from the analysis that BERT is the best technique to use in the case of domain specific data.


\keywords{Semantic similarity  \and Sentence embedding \and language representation model \and Universal sentence encoder \and InferSent \and BERT}
\end{abstract}
\section{Introduction}
Semantic similarity estimation of text is an important and open research problem in natural language processing as well as in natural language understanding \cite{ref_young}. Two texts are semantically similar if they mean the same thing \cite{ref_semeval}. And this semantic similarity estimation lays foundation to various downstream applications including information retrieval, document clustering, word-sense disambiguation, machine translation, text summarization, text generation, question answering (QA), short answer grading, automatic essay scoring, semantic search, dialog and conversational systems \cite{ref_semeval}. Realizing the potential of this area, the Semantic Textual Similarity (STS) shared task has been held annually since 2012, providing a venue for evaluation of state-of- the-art algorithms and models for semantic similarity estimation \cite{ref_semeval_2012,ref_semeval_2013,ref_semeval_2014,ref_semeval_2015,ref_semeval_2016}. 

In the application area of question answering, generally a knowledge base (KB) is maintained containing queries and associated answers \cite{ref_young}. As the same query can be asked in many different ways, it becomes important to find the most semantically similar query from the KB to recommend an answer. For instance, the answers for "To buy stocks what's the minimum" and "what is the minimum amount to buy a stock" is same. And question answering is one of the key research areas at Fidelity. 

There are many research articles covering textual semantic similarity estimation~\cite{ref_use,ref_infersent,ref_bert,ref_perone}. Word embeddings such as word2vec \cite{ref_w2v}, Glove \cite{ref_glove} and FastText \cite{ref_fasttext1} have been used widely in NLP to compute  vector representation for words capturing the semantics that can be used to compute the semantic similarity between words. Although word embeddings effectively capture the semantics for words, they are more prone to yield limited performance when used for sentence embedding via averaging of word vectors or via some other way \cite{ref_perone,ref_use,ref_infersent}. Reasons for this decreased performance could be because, usually word embeddings do not consider the word order into account and the word embeddings are learnt in an unsupervised manner. 

Due to these shortcomings, recent research has focused on embedding of larger text, such as sentences, paragraphs or documents, straightaway, rather than looking at word embeddings. A lot of these research has lead to the development of various pre-trained models which includes Google's Universal Sentence Encoder (USE) \cite{ref_use} and Facebook's InferSent \cite{ref_infersent}. Generally, these pre-trained models have been trained on both supervised as well as unsupervised tasks in order to capture as much universal semantic information as possible. And these models have shown better performance on various downstream tasks (as well as on transfer learning) as compared to word embeddings. Along with research on sentence embedding, there is a big focus in the area of language model pretraining which has shown to give improved performance in various NLP tasks including predicting the relationship between sentences \cite{ref_elmo,ref_openai}. Very recently, Google had open sourced a new language representation model called BERT \cite{ref_bert} which stands for Bidirectional Encoder Representations from Transformers, and it has shown to outperform other techniques in many NLP tasks such as text classification, textual entailment, question answering and language inference, without substantial task-specific architecture modifications, easily making it among the best in the state-of-the-art methods.

In this paper, we use sentence embedding pre-trained models including Universal sentence encoder and InferSent, and language representation model of BERT for estimating the semantic similarity for pairs of text inputs. Along with this, we also use string matching method using the Ratcliff/Obershelp pattern-matching algorithm \cite{ref_ratcliff} which acts as the baseline for our analysis. For the experiments, we use an in-house dataset and a  public dataset from Quora, called the Quora Question Pairs (QQP) \cite{ref_qqp}. 

\section{Materials}

\subsection{Dataset details}

\subsubsection{Quora dataset: }
The Quora dataset consist of question pairs (that were asked on Quora by its users) along with a label indicating semantic equivalence. There are 404290 question pairs where 149263 (37\%) were similar and the rest 255027 (63\%) were not similar. This dataset was general in nature, capturing the questions asked by the Quora users that were not specific to any domain. The dataset was pre-processed and after pre-possessing, three observations were removed that belonged to the positive class.

 \subsubsection{Fidelity data: }
The Fidelity data set consist of user query pairs along with the label for similarity. There are 3757 query pairs where 1703 (45\%) are similar and the rest 2054 (55\%) are not similar. This data is domain specific and it contained questions pertinent to the financial services domain.

\section{Methods}
\begin{figure}[h]
  \includegraphics[width=\linewidth]{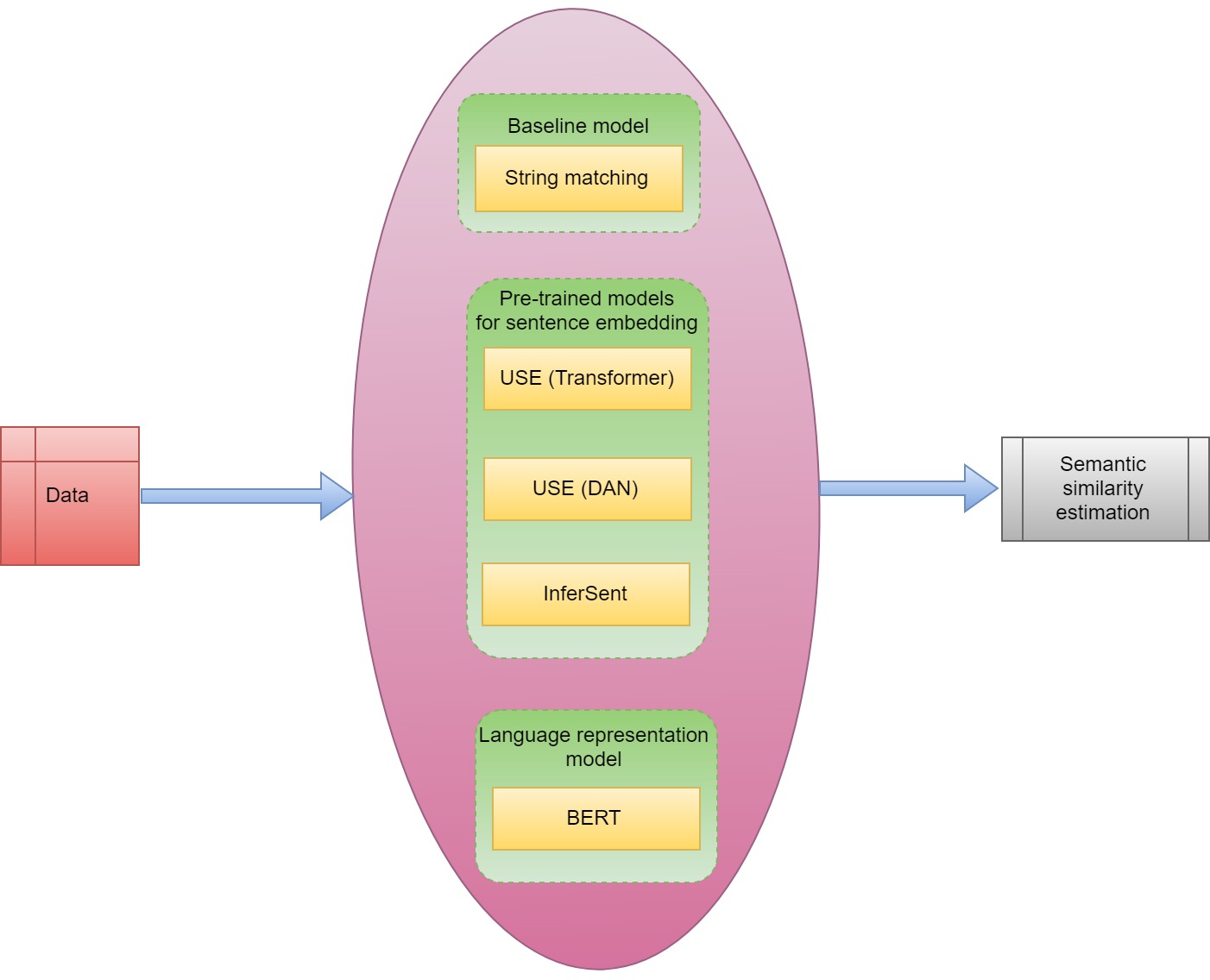}
  \caption{A high level flow chart of the textual similarity estimation process }
  \label{fig:flow_chart}
\end{figure}

A high level overview of the analysis carried out in the paper is as shown in Fig. \ref{fig:flow_chart}. As mentioned above, we use the Quora dataset and an in-house dataset as well for the experiments. The purpose of using Quora dataset is for the sake of comparison with a public dataset. BERT has already been applied on the QQP dataset and have shown to beat state-of-the-art results.

The datasets are given as inputs to different techniques, starting with string matching using the Ratcliff/Obershelp algorithm which effectively acts as a baseline model, pre-trained models including USE and InferSent for embedding text, and also the language representation model of BERT. 

The string matching method using the Ratcliff/Obershelp algorithm basically measures the similarity of sequences based on the number of matching characters. To account for the order of words in the sequence, the sequence is tokenized followed by sorting the tokens alphabetically, and then joining them back into a string. Before the string matching procedure, the sequences are lemmatized for the purpose of normalization.

The pre-trained models (USE and InferSent) for text embedding generate embeddings for the sequences or the input question pairs and then appropriate metric is used to find dissimilarity or distance between the two embeddings to estimate the semantic similarity. The authors of USE state in their paper \cite{ref_use} that to find the similarity of the sentence embeddings, they first compute the cosine similarity of the two embeddings and then use arccos to convert the cosine similarity into an angular distance. They observed that using a similarity based on angular distance performs better on average than raw cosine similarity. However, in the InferSent paper \cite{ref_infersent}, they stick with cosine distance between the embedding vectors. In this paper, we use the angular distance for the USE embeddings and cosine distance for the InferSent embeddings.

The BERT model which is basically a language representation model pre-trains deep bidirectional representations by jointly conditioning on both left and right context in all layers. And due to this, the pre-trained BERT representations can be fine-tuned with just one additional output layer to create state-of-the-art models for a wide range of tasks such as sentence classification.

\subsection{String matching using Ratcliff/Obershelp algorithm}
The Ratcliff/Obershelp algorithm computes the similarity of two text as the doubled number of matching characters divided by the total number of characters in the two strings \cite{ref_ratcliff}. Matching characters are the ones in the longest common subsequence plus, recursively, matching characters in the unmatched region on either side of the longest common subsequence. It returns a score between 0 for no match and 100 for an all match.

\subsection{Universal Sentence Encoder (USE)}
In early 2018, Google developed USE \cite{ref_use} to encode sentences to embedding vectors. And they shared two variants of the encoding model. One used deep averaging network (DAN) and the other was based on the transformer architecture. 

\subsubsection{USE (Transformer) ---} 
The transformer based encoding model generates sentence embeddings using the encoding sub-graph of the transformer architecture. This sub-graph uses attention to create context aware representations of words in a sentence, taking both the order and identity of all the other words into consideration \cite{ref_transformer}. These representations are then converted to a fixed length sentence encoding vector (512 dimensional) by computing the element-wise sum of the representations at each word position followed by dividing by the square root of the length of the sentence so that the differences between short sentences are not dominated by sentence length effects \cite{ref_use}. 

To make the encoding model as general as possible, a multi-task learning was carried out where the encoding model was used to feed multiple downstream tasks. The tasks include a Skip-Thought like task for the unsupervised learning from arbitrary  text \cite{ref_skip}; a conversational input-response task \cite{ref_henderson}; and classification tasks for training on supervised data. For word level transfer, word embeddings from a word2vec skip-gram model \cite{ref_w2v} trained on a corpus of news data was used. 

\subsubsection{USE (DAN) ---} 
This encoding model makes use of a DAN where input embeddings for words and bi-grams are first averaged together and then passed through a feedforward deep neural network (DNN) to produce sentence embeddings \cite{ref_use,ref_dan}. Similar to the transformer based model, the DAN based USE is trained via multi-task learning and it also outputs a 512 dimensional vector.

\subsection{InferSent}
InferSent \cite{ref_infersent} is a sentence embedding pre-trained model that was developed by Facebook AI Research. It is an encoder model based on a bi-directional LSTM architecture with max pooling, trained on the supervised data of Stanford Natural Language Inference (SNLI) \cite{ref_slni} datasets. The architecture consist of a shared sentence encoder which is based on bi-directional LSTM and outputs a representation for the premise $u$ and the hypothesis $v$. After the vectors are generated, a new vector is generated by combining them in 3 ways, (i) concatenation of the two representations $(u, v)$; (ii) element-wise product $u \ast v$; and (iii) absolute element-wise difference $\mid u-v\mid$. This resulting vector, which captures information from both the premise and the hypothesis, is fed into a 3-class classifier consisting of fully connected layers culminating in a softmax layer. 

For training this model, stochastic gradient descent with learning rate of 0.1 and a weight decay of 0.99 were used. This learning rate was divided by 5 if the dev accuracy decreased. Mini-batch size of 64 was used and the training was stoped when the learning rate goes under the threshold of $10^{-5}$.  For the classifier, a multi-layer perceptron with 1 hidden-layer of 512 hidden units was used. 

Facebook open-sourced two versions of the pre-trained model: V1 that was trained using GloVe \cite{ref_glove}, and V2 that was trained using fastText \cite{ref_fasttext1}. Although both the models are competent, for the semantic similarity task, the V2 version was performing better. In this paper, we use the V2 version.

\subsection{BERT}
By the end of 2018, Google open-sourced BERT \cite{ref_bert} which is a new language representation model that is a multi-layer bidirectional Transformer encoder based on the Transformer architecture as in \cite{ref_transformer}. The BERT model is pretrained using two unsupervised prediction tasks. One is the "masked language model" where some of the tokens (around 15\%) were randomly masked, and the objective is to predict the masked word based on only its context. The second task is that of "next sentence prediction" where the aim is to predict the next sentence. The training loss of the model which is optimized, is the sum of mean masked language model likelihood and mean next sentence prediction likelihood. 

For the pretraining corpus, BooksCorpus (800M words) \cite{ref_bookcorpus} and English Wikipedia (2,500M words) were used. Batch size used was 256 sequences with each sequence containing 512 tokens, training was done for 1,000,000 steps or 40 epochs over the 3.3 billion word corpus. Adam optimizer with learning rate of 1e-4, $\beta1 = 0.9$, $\beta2 = 0.999$, L2 weight decay of 0.01, learning rate warmup over the first 10,000 steps, and linear decay of the learning rate were used. Activation function used was Gaussian Error Linear Unit (GELU) \cite{ref_gelu} and dropout probability of 0.1 was used on all layers. They develop two variants of models based on the size. One is $BERT_{BASE}$ with number of layers (Transformer blocks) as 12, the hidden size as 768, the number of self-attention heads as 12, and total number of parameters as 110M. The other is $BERT_{LARGE}$ which has 340M parameters. In the paper, we use the $BERT_{BASE}$ model due to infrastructure constraints. 

\section{Results and Discussion}
The performance measures on the two datasets for different methods used in the paper are shown in Table \ref{tab1}. To get optimal results based on the Region Operating Characteristic (ROC) curve, we use the Youden's index \cite{ref_youden1} to obtain an optimal cut-off point in the ROC \cite{ref_youden2}. The results reported for BERT for the Quora dataset in the table are taken from the original paper for BERT \cite{ref_bert}, and the original paper reports only the accuracy. F-score was taken from the GLUE leaderboard (https://gluebenchmark.com/leaderboard). They do not report other metrics which are  sensitivity, specificity, precision and the area under the ROC (AUC). BERT gave best results for both the datasets. 

For the Quora, dataset, string matching provided a decent baseline result which was 64.5\% accuracy and 72.7\% AUC. Among the sentence embedding pre-trained models, USE (Transformer) provided the best results, better than USE (DAN) and InferSent. This should be because of its transformer based architecture which specifically targets high accuracy at the cost of greater model complexity and resource consumption. 

For the Fidelity dataset, it is observed that the pre-trained models performed much inferior to the string matching based method. This might be because this dataset has a majority of user questions which are not straightforward, as this dataset contains different ways of asking the same question in different ways. Another reason could be that a majority of the questions are highly domain dependent which caused the decreased performance from the pre-trained sentence encoder models which are developed based on publicly available data that are general in nature and not specific to any domain. BERT showed a tremendous improvement over the other methods. This could be due to the fine-tuning step that is involved in the BERT training process which makes use of a training data allowing it to fine-tune based on the training data. 

It is to be noted that we have used $BERT_{BASE}$ , and not $BERT_{LARGE}$ for our experiments due to restrictions in infrastructure. The original BERT paper \cite{ref_bert} reports that the $BERT_{LARGE}$ model significantly outperforms $BERT_{BASE}$ across many tasks. Using $BERT_{LARGE}$ may further improve the result. We also tried different values for the number of epochs for training the BERT model (Table \ref{tab4}). We tried with 3, 5 and 10, and observed that 5 epochs gave the best result.

\subsection{Note on the resource usage}
Among the methods, the string matching technique was the fastest. In its worst-case scenario, when there is no match of characters between two sequences, $N\times M$ number of comparisons are required, where N is the number of characters in the first string and M is the number of characters in the second string. It takes very little memory for computation. For USE (DAN), the time complexity is $O(n)$ and memory requirement is constant in sentence length. However, for USE (Transformer),  the time complexity is $O(n^2)$ and space complexity also scales quadratically, $O(n^2)$ in sentence length. Comparing USE (DAN) and USE (Transformer), the former is remarkably computational efficient, but at the cost of slight decreased performance than the latter,  as the compute time for the DAN model stays nearly constant as sentence length is increased. BERT models are enormous models with millions of parameters. But the good thing is that the pre-training was already carried out, and we just had to focus on fine tuning the BERT pre-trained models. As per the official GitHub page for BERT (https://github.com/google-research/bert), $BERT_{BASE}$  could run on a GPU that has at least 12GB of RAM using the hyperparameters specified. In the case of $BERT_{LARGE}$, they recommend a Cloud TPU, which has 64GB of RAM. And they also mention that running $BERT_{LARGE}$ on a GPU with 12GB - 16GB of RAM is likely to give out-of-memory issues. 

\begin{table}[h]
\setlength{\tabcolsep}{5pt}
\renewcommand{\arraystretch}{1.4}
\caption{Performance measures on the different datasets}\label{tab1}
\begin{tabular}{|c|c|c|c|c|c|c|}
\hline
  \multicolumn{7}{|c|}{\textbf{\textit{Quora dataset}}} \\
\hline
\it{Method} & \it{Accuracy} & \it{Sensitivity} & \it{Specificity} & \it{AUC} & \it{Precision} & \it{F-score}\\
\hline
String matching & 64.5 & 81.9 & 54.4 & 72.7 & 51.2 & 63.0\\
USE (DAN) & 70.9 & 83.5 & 63.6 & 79.9 & 57.3 & 68.0\\
USE (Transformer) & 73.9 & 82.9 & 68.6 & 82.7 & 60.7 & 70.1\\
InferSent & 67.9 & 84.4 & 58.2 & 75.9 & 54.2 & 66.0 \\
\bf{BERT} \footnote{This is a footnote.}&  \bf{72.1} & -- & -- & -- & -- & \bf{89.3}\\
\hline
  \multicolumn{7}{|c|}{\textbf{\textit{Fidelity dataset} }}\\
\hline
\it{Method} & \it{Accuracy} & \it{Sensitivity} & \it{Specificity} & \it{AUC} & \it{Precision} & \it{F-score}\\
\hline
String matching & 67.6 & 71.1 & 64.6 & 75.3 & 62.5 & 67.0\\
USE (DAN) & 63.4 & 66.6 & 60.8 & 68.5 & 58.5 & 62.0\\
USE (Transformer) & 67.7 & 68.3 & 67.2 & 72.9 & 63.3 & 65.7\\
InferSent & 65.5 & 60.7 & 69.5 & 70.8 & 62.3 & 61.0\\
\bf{BERT} & \bf{80.3} & \bf{80.6} & \bf{80.0} & \bf{88.4} & \bf{77.0}  & \bf{78.8}\\
\hline
\multicolumn{7}{p{\textwidth}}{USE (DAN) and USE (Transformer) are DAN based and transformer based USE, respectively. $^{1}$ The performance measures for BERT on the Quora dataset have not been computed in the paper and is as reported in the original paper for BERT \cite{ref_bert}}.\\
\end{tabular}
\end{table}

\begin{table}[h]
\setlength{\tabcolsep}{8.8pt}
\renewcommand{\arraystretch}{1.4}
\caption{Performances of BERT with different epochs}\label{tab4}
\begin{tabular}{|c|c|c|c|c|c|c|}
\hline
\it{epochs} & \it{Accuracy} & \it{Sensitivity} & \it{Specificity} & \it{AUC} & \it{Precision} & \it{F-score}\\
\hline
3 & 79.2 & 80.9 & 77.8 & 88.2 & 75.1  & 77.9\\
\bf{5} & \bf{80.3} & 80.6 & \bf{80.0} & \bf{88.4} & \bf{77.0}  & \bf{78.8}\\
10 & 80.1 & \bf{81.4} & 78.9 & 87.6 & 76.2  & 78.7\\
\hline
\end{tabular}
\end{table}

\subsection{Note on misclassified instances}
Table \ref{tab2} shows few misclassified instances from the BERT method. The first five examples show cases which were misclassified as semantically similar. This might be because that these examples had almost similar structure making the classifier to predict that they are semantically equivalent. On the other hand, the rest of the examples were misclassified as semantically different. These were few user query pairs which are semantically same but asked in two very different ways. For example, in the question pair "\textit{are there maintenance fees for stocks}" and "\textit{Are there any management fees to own the stock?}", the BERT model could not detect the semantic equivalence as it might have failed to estimate the similarity between maintenance fees and management fees. One reason for this might be because of the limited data that was used for training. Having a bigger dataset might essentially rectify this issue. We could observe that there were few spelling errors (we did not carry out spelling correction in this work) in these examples, such as accouont, fideliy etc. These might have added to the cause for    misclassification.

\begin{table}[h]
\setlength{\tabcolsep}{5pt}
\renewcommand{\arraystretch}{1.4}
\caption{Examples of misclassified instances using BERT}\label{tab2}
\begin{tabular}{|p{5cm}|p{5cm}|p{0.5cm}|p{0.5cm}|}
\hline
\hfil \it{Qn1} & \hfil \it{Qn2} & \it{true} & \it{pred}\\
\hline
Do I need to get active trader account to trade penny stocks? & Do you trade all penny stocks? & 0 & 1\\
Why are the reinvested dividends not shown? & Can you please show me how to reinvest dividends myself? & 0 & 1\\
Would you do a 'limit' market order? & Would I do market order or limit order & 0 & 1\\
How do I find out which stocks are easy to borrow? & How do I know a stock is hard to borrow? & 0 & 1\\
Can you retroactively reinvest dividends? & Can I set up account to reinvest dividends? & 0 & 1\\
Can I open a brokerage accouont in my existing ira & Can I open an ira brokerage account? & 1 & 0\\
Does IRA fall under brokerage & Does an IRA account = a Fidelity brokerage account? & 1 & 0\\
Are there maintenance fees for stocks & Are there any management fees to own the stock? & 1 & 0\\
Can I buy mutual funds beyond Fidelity's own funds? & Can I buy into non fideliy mutual funds? & 1 & 0\\
How do I protect myself on mutual funds? & I would like to discuss how safe mutual funds are & 1 & 0\\
When I sell shares how is the price calculated & When I sell the stock how do i know what price i'm selling at? & 1 & 0\\
Can I short a mutual fund? & Can I short mutual funds? & 1 & 0\\
\hline
\multicolumn{4}{p{\textwidth}}{\textit{true} is the actual label and \textit{pred} is the predicted label}\\
\end{tabular}
\end{table}

\section{Conclusion}
In this work, we estimate the semantic similarity of question pairs from an in-house dataset as well as a public dataset using state-of-the-art techniques. We observe that string matching provided a decent performance although it cannot detect semantic equivalence. Among the pretrained sentence embedding models, USE (Transformer) performed better than USE(DAN) and InferSent although the computation time was high. Overall, BERT method gave the best results among all the techniques used. This should be due to the fine-tuning that is involved in the training process allowing it to learn based on the training data. We demonstrate in this work the applicability of BERT on domain specific dataset like ours.

\section*{Acknowledgement}
Sincere thanks to Bibhash Chakrabarty and Manish Gupta for sharing their valuable insights that helped in the paper. Thanks to Manish Gupta for providing and sharing the Fidelity dataset. Special thanks to Debasis Bal without whose support this paper would not have come to life.

 \bibliographystyle{splncs04}
%

\end{document}